\documentclass[11pt,a4paper]{article}
\usepackage[hyperref]{naaclhlt2018}
\usepackage{times}
\usepackage{latexsym}

\aclfinalcopy
\usepackage{CJK}
\usepackage{amsmath}
\usepackage{amssymb}

\DeclareMathOperator*{\softmax}{softmax}

\usepackage{multirow}
\usepackage{times}
\usepackage{graphics}
\usepackage{amsfonts}
\usepackage{graphicx}
\usepackage{makecell}
\usepackage{array}
\usepackage{bbding}
\usepackage{color,framed}

\usepackage[ruled,vlined,linesnumbered]{algorithm2e}
\usepackage{booktabs}  

\usepackage{tikz}

\title{Interpretable Charge Predictions for Criminal Cases:\\Learning to Generate Court Views from Fact Descriptions}
\author{$^1$Hai Ye$^\ast$, $^1$Xin Jiang$^\ast$, $^2$Zhunchen Luo\thanks{\ \ \ indicates equal contribution.} , $^1$Wenhan Chao\thanks{\ \ \ Corresponding author.}\\ 
$^1$ School of Computer Science and Engineering, Beihang University\\
$^2$ Information Research Center of Military Science\\PLA Academy of Military Science\\
$^{1, 2}$ Beijing, China\\
$^1$ \{yehai, xinjiang, chaowenhan\}@buaa.edu.cn\\$^2$ zhunchenluo@gmail.com
}
\date{}

\begin{document}
\begin{CJK*}{UTF8}{gbsn}
\maketitle

\begin{abstract}
In this paper, we propose to study the problem of $\text{\scshape{Court}}$ $\text{\scshape{View}}$ $\text{\scshape{Gen}}$eration from the fact description in a criminal case. The task aims to improve the interpretability of charge prediction systems and help automatic legal document generation. We formulate this task as a text-to-text natural language generation (NLG) problem. Sequence-to-sequence model has achieved cutting-edge performances in many NLG tasks. However, due to the non-distinctions of fact descriptions, it is hard for Seq2Seq model to generate charge-discriminative court views. In this work, we explore charge labels to tackle this issue. 
We propose a label-conditioned Seq2Seq model with attention for this problem, to decode court views conditioned on encoded charge labels. 
Experimental results show the effectiveness of our method.\footnote{Data and codes are available at \url{https://github.com/oceanypt/Court-View-Gen}.} 
\end{abstract}

\section{Introduction}
Previous work has brought up multiple legal assistant systems with various functions, such as finding relevant cases given the query \cite{relevant-case-chen}, providing applicable law articles for a given case \cite{article-liu-1} and etc., which have substantially improved the working efficiency. As legal assistant systems, charge prediction systems aim to determine appropriate charges such as \emph{homicide} and \emph{assault} for varied criminal cases by analyzing textual fact descriptions from cases \cite{luo2017}, but ignore to give out the interpretations for the charge determination. 

\begin{figure*} 
\centering\includegraphics[width = \textwidth]{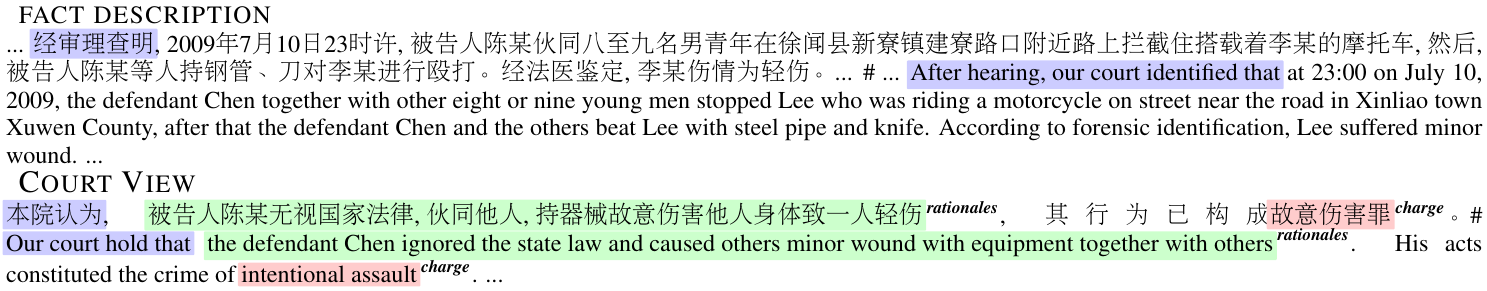}
\caption{An example of fact description and court view from a legal document in a case.}
\label{court-view}
\end{figure*}

\emph{Court view} is the written explanation from judges to interprete the charge decision for certain criminal case and is also the core part in a legal document, which consists of \emph{rationales} and a \emph{charge} where the charge is supported by the rationales as shown in Fig. \ref{court-view}. In this work, we propose to study the problem of $\text{\scshape{Court}}$ $\text{\scshape{View}}$ $\text{\scshape{Gen}}$eration from fact descriptions in cases, and we formulate it as a text-to-text natural language generation (NLG) problem \cite{NLGSurvey}. 
The input is the fact description in a case and the output is the corresponding court view. We only focus on generating rationales because charges can be decided by judges or charge prediction systems by also analyzing the fact descriptions \cite{luo2017, charge-Lin2012}. 
$\text{\scshape{Court-View-Gen}}$ has beneficial functions, in that:
(1) improve the interpretability of charge prediction systems by generating rationales in court views to support the predicted charges. The justification for charge decision is as important as deciding the charge itself \cite{visual-exp, lei}. 
(2) benefit the automatic legal document generation as legal assistant systems, by automatically generating court views from fact descriptions, to release much human labor especially for simple cases but in large amount, where fact descriptions can be obtained from legal professionals or techniques such as information extraction \cite{IE}. 



$\text{\scshape{Court-View-Gen}}$ is not a trivial task. High-quality rationales in court views should contain the important fact details such as the degree of injury for charge of \emph{intentional injury}, as they are important basis for charge determination. Fact details are like the summary for the fact description similar to the task of $\text{\scshape{doc}}$ument $\text{\scshape{sum}}$marization \cite{DBLP:journals/kais/YaoWX17}. However, rationales are not the simple summary with only fact details, to support charges, they should be charge-discriminative with \emph{deduced information} which does not appear in fact descriptions. The fact descriptions for charge of \emph{negligent homicide} usually only describe someone being killed without direct statement about the motive for killing, $\text{\scshape{doc-sum}}$ will only summarize the fact of someone being killed, but rationales have to further contain the killing intention, aiming to be discriminative from those rationales for other charges like \emph{intentional homicide}. However, it is hard to generate charge-discriminative rationales when input fact descriptions are not distinct among other facts with different charges. The fact descriptions for charge of \emph{intentional homicide} are similar to those for charge of \emph{negligent homicide} and also describe someone being killed but without clear motive, making it hard to generate charge-discriminative court views with accurate killing motives among the two charges.    
Recently, sequence-to-sequence model with encoder-decoder paradigm \cite{Seq2SeqNN} has achieved cutting-edge results in many NLG tasks, such as paraphrase \cite{LapataSM17}, code generation \cite{code:ling} and question generation \cite{Q-Gen}. Seq2Seq model has also exhibited state-of-the-art performances on task of $\text{\scshape{doc-sum}}$ \cite{Sum-Chopra, Sum-Wan}. 
However, non-distinctions of fact descriptions render Seq2Seq model hard to generate charge-discriminative rationales. In this paper, we explore charge labels of the corresponding fact descriptions, to benefit generating charge-discriminative rationales, where charge labels can be easily decided by human or charge prediction systems. Charge labels will provide with extra information to classify the non-discriminative fact descriptions. We propose a \emph{label-conditioned} Seq2Seq model with attention for our task, in which 
fact descriptions are encoded into context vectors by an encoder and a decoder generates rationales with these vectors. 
We further encode charges as the labels and decode the rationales conditioned on the labels, to entail the decoder to learn to select gold-charge-related words to decode. 
Widely used attention mechanism \cite{Luong-att} is fused into the Seq2Seq model, to learn to align target words to fact details in fact descriptions. 
Similar to \citet{luo2017}, we evaluate our model on Chinese criminal cases by constructing dataset from Chinese government website. 

Our contributions in this paper can be summarized as follows:

$\bullet$ We propose the task of \emph{court view generation} and release a real-world dataset for this task.


$\bullet$ We formulate the task as a text-to-text NLG problem. We utilize charge labels to benefit charge-discriminative court views generation, and propose a label-conditioned sequence-to-sequence model with attention for this task. 

$\bullet$ Extensive experiments are conducted on a real-world dataset. The results show the efficiency of our model and exploiting charge labels for charge-discriminations improvement.

\section{Related Work}
Our work is firstly related to previous studies on legal assistant systems. Previous work considers the task of charge prediction as a text classification problem \cite{luo2017, charge-Liu2004, charge-Liu2006, charge-Lin2012}. Recently, \citet{luo2017} investigate deep learning methods for this task. Besides, there are also works on identifying applicable articles for a given case \cite{article-liu-1, charge-Liu2006, article-Liu2015}, answering legal questions as a consulting system \cite{legal-qa-Kim, legal-qa-2} and searching relevant cases for a given query \cite{relevant-law-1, relevant-case-chen}. As a legal assistant system, $\text{\scshape{Court-View-Gen}}$ can benefit automatic legal document generation by generating court views from fact descriptions obtained from the last phase, through legal professionals or other technics like information extraction \cite{IE} from raw documents in a case, if we generate legal documents step by step. 

Our work is also related to recent studies on model interpretation \cite{WhyTrustYou, Mythos, Program}. 
Recently, much work has paid attention to giving textual explanations for classifications. \citet{visual-exp} generate visual explanations for image classification. \citet{lei} propose to learn to select most supportive snippets from raw texts for text classification. 
$\text{\scshape{Court-}}$$\text{\scshape{View-}}$$\text{\scshape{Gen}}$ can improve the interpretability of charge prediction systems by generating textual court views when predict the charges.

Our label-conditioned Seq2Seq model steams from widely used encoder-decoder paradigm \cite{Seq2SeqNN} which has been widely used in machine translation \cite{NMTAT, Luong-att}, summarization \cite{Sum-Wan, Sum-CoNLL,Sum-Chopra,Sum-Lapata}, semantic parsing \cite{dong-lapata:2016:P16-1} and paraphrase \cite{LapataSM17} or other NLG problems such as product review generation \cite{Product-gen} and code generation \cite{code:yin, code:ling}. 
\citet{visual-exp} propose to encode image labels for visual-language models to generate justification texts for image classification. 
We also introduce charge labels into Seq2Seq model to improve the charge-discriminations of generated rationales. Widely used attention mechanism \cite{Luong-att, DBLP:conf/icml/2015} is applied to generate fact details more accurately.

\section{$\text{\scshape{Court-View-Gen}}$ Problem} 
\emph{\textbf{Court View}} is the judicial explanation to interpret the reasons for the court making such charge for a case, consisting of the rationales and the charge supported by the rationales as shown in Fig. \ref{court-view}. In this work, we only focus on generating the part of rationales in court views. Charge prediction can be achieved by human or charge prediction systems \cite{luo2017}. Final court views can be easily constructed by combining the generated rationales and the pre-decided charges.
 
\noindent{\emph{\textbf{Fact Description}}} is the identified facts in a case (relevant events that have happened) such as the criminal acts (e.g. \emph{degree of injury}). 
 
 The input of our model is the word sequential fact description in a case and the output is a word sequential court view (rationales part). We define the fact description as $\mathbf{\mathrm{x}} = (x_1, x_2, \cdots, x_{|\mathrm{x}|})$ and the corresponding rationales as $\mathrm{y} = (y_1, y_2, \cdots, y_{|\mathrm{y}|})$. The charge for the case is denoted as $\mathrm{v}$ and will be exploited for $\text{\scshape{Court-View-Gen}}$. The task of $\text{\scshape{Court-View-Gen}}$ is to find $\hat{\mathrm{y}}$ given $\mathrm{x}$ conditioned on the charge label $\mathrm{v}$:
 \begin{equation}
 \hat{\mathrm{y}} = \arg \max_{\mathrm{y}} P(\mathrm{y}|\mathrm{x}, \mathrm{v})
 \label{all-define}
 \end{equation}
 where $P(\mathrm{y}|\mathrm{x}, \mathrm{v})$ is the likelihood of the predicted rationales in the court view.

\section{Our Model}
\subsection{Sequence-to-Sequence Model with Attention}
\begin{figure}
\centering\includegraphics[width = \columnwidth]{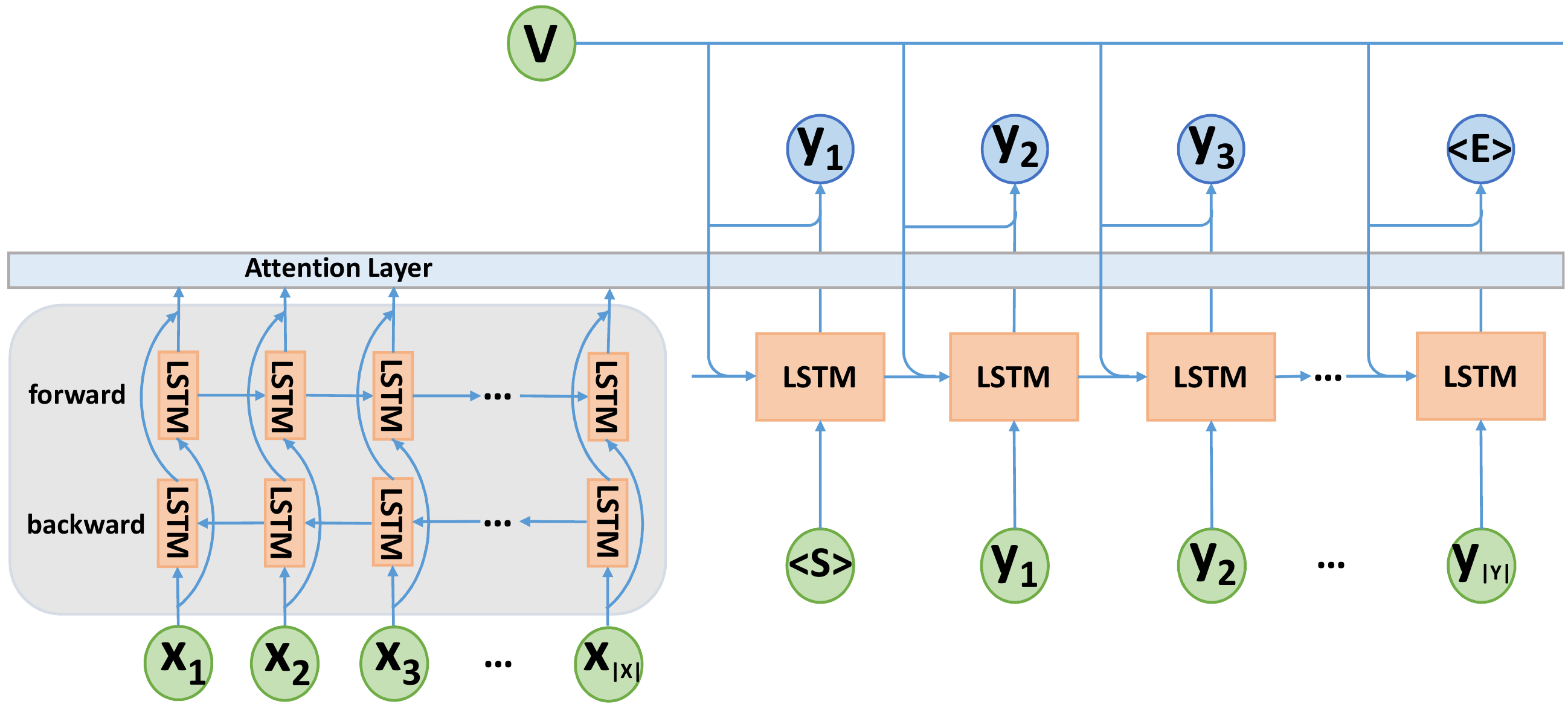}
\caption{\small{Label-conditioned Seq2Seq model with attention.}}
\label{model}
\end{figure}

Similar to \citet{Luong-att}, our Seq2Seq model consists of an encoder and a decoder as shown in Fig. \ref{model}. Given the pair of fact description and rationales in court view ($\mathbf{\mathrm{x}}, \mathbf{\mathrm{y}}$), the encoder reads the word sequence of $\mathbf{\mathrm{x}}$ and then the decoder will learn to predict the rationales in court view $\mathrm{y}$. The probability of predicted $\mathrm{y}$ is given as follows:
\begin{equation}
 P(\mathrm{y}) = \prod_{i=1}^{|\mathrm{y}|} P(y_i|y_{<i}, \mathrm{x}) \label{f-base}
\end{equation}
where $y_{<i} = y_1, y_2, \cdots, y_{i-1}$. We use a bidirectional LSTM \cite{lstm} as encoder and use another LSTM as decoder similar to \citet{Q-Gen}. 

\noindent{\textbf{Decoder.}} From the decoder side, at time $t$, the probability to predict $y_t$ is computed as follows:
\begin{equation}
\nonumber P(y_t | y_{<t}, \mathbf{c}_t) = \softmax(\mathbf{W}_1\tanh( \mathbf{W}_0[\mathbf{s}_t;\mathbf{c}_t]))
\end{equation}
where $\mathbf{W}_0$ and $\mathbf{W}_1$ are learnable parameters; $\mathbf{s}_t$ is the hidden state of decoder at time $t$; $\mathbf{c}_t$ is the context vector generated from the encoder side containing the information of $\mathrm{x}$ at time $t$; here the bias of model is omitted for simplification. The hidden state of $\mathbf{s}_t$ is computed as follows:
\begin{equation}
\nonumber \mathbf{s}_t = \mathrm{LSTM}_d(y_{t-1},\mathbf{s}_{t-1})
\end{equation}
where $y_{t-1}$ is the word embedding vector for pre-state target word at time $t-1$. The initial state for decoder is initialized by the last state of encoder. 

Context vector of $\mathbf{c}_t$ is computed by summing up the hidden states of $\{\mathbf{h}_k\}^{|\mathrm{x}|}_{k=1}$ generated by the encoder with attention mechanism and we adopt global attention \cite{Luong-att} in our work. 

\noindent{\textbf{Encoder with Attention.}}
We adopt a one-layer bidirectional LSTM to encoder the fact descriptions. The hidden state $\mathbf{h}_j$ at time $j$ is computed as follows:
\begin{equation}
\nonumber \mathbf{h}_j = [\overrightarrow{\mathbf{h}_j}; \overleftarrow{\mathbf{h}_j}]
\end{equation}
where $\mathbf{h}_j$ is the concatenation of forward hidden state $\nonumber \overrightarrow{\mathbf{h}_j}$ and backward hidden state $\nonumber  \overleftarrow{\mathbf{h}_j}$, specifically:
\begin{align}
\nonumber \overrightarrow{\mathbf{h}_j} & = \overrightarrow{ \mathrm{LSTM}}_e(x_j, \overrightarrow{ \mathbf{h}}_{j-1}) \\
\nonumber  \overleftarrow{\mathbf{h}_j} & = \overleftarrow{ \mathrm{LSTM}}_e(x_j, \overleftarrow{ \mathbf{h}}_{j+1}) 
\end{align}
The hidden outputs $\{\mathbf{h}_k\}_{k=1}^{|\mathrm{x}|}$ will be used to compute the context vectors for decoder.  

From the decoder side, by applying attention mechanism at time $i$, the context vector of $\mathbf{c}_i$ is generated as follows:
 \begin{equation}
 \mathbf{c}_i = \sum_{j = 1}^{|{ \mathrm{x}}|}   \alpha_{ij} \mathbf{h}_j
 \label{att-1}
\end{equation}
where $\alpha_{ij}$ is the attention weight and is computed as follows:
\begin{align}
\alpha_{ij} & = \frac{ \exp(\mathbf{s}_{i}^T\mathbf{W}_2\mathbf{h}_j) }{ \sum_{k=1}^{|{\mathrm{x}}|}  \exp(\mathbf{s}_{i}^T\mathbf{W}_2\mathbf{h}_k)} 
\label{att-2}
\end{align}
where $\mathbf{s}_i$ is the hidden output state at time $i$ in the decoder side.

\subsection{Label-conditioned Sequence-to-Sequence Model with Attention}
Given the tuple of fact description, rationales in court view and charge label ($\mathrm{x}, \mathrm{y}, \mathrm{v}$), the probability to predict $\mathrm{y}$ is computed as follows: 
\begin{equation}
 P(\mathrm{y}) = \prod_{i=1}^{|\mathrm{y}|} P(y_i|y_{<i}, \mathrm{x}, \mathrm{v}) 
\end{equation}
From this formula, encoding charge labels provides extra constrains comparing to Eq. (\ref{f-base}), and restricts the target word searching space from the whole space to only gold-charge-related space for rationales generation, so model can generate more charge-distinct rationales. 
Charge labels are trainable parameters denoted by $\mathbf{E}^{{v}}$ where every charge will have a trainable vector from $\mathbf{E}^{{v}}$, which will be updated in the model training process.

As shown in Fig. \ref{model}, in the decoder side, at time $t$, $y_t$ is predicted with the probability as follows:
\begin{align}
\nonumber P(y_t|y_{<t}, \mathbf{c}_t, \mathrm{v}) & = \\ 
\softmax(\mathbf{W}_1 & \tanh(\mathbf{W}_0[\mathbf{s}_t;\mathbf{c}_t;\mathbf{E}^{{v}}_{[\mathrm{v}]}])) 
\label{dict}
\end{align}
where $\mathbf{E}^{{v}}_{[\mathrm{v}]}$ is the embedding vector of $\mathrm{v}$ obtained from $\mathbf{E}^{{v}}$. In this formula, we connect charge label $\mathrm{v}$ to $\mathbf{s}_t$ and $\mathbf{c}_t$ aiming to influence the word selection process. We hope that our model can learn the latent connections between the charge label $\mathrm{v}$ and the words of rationales in court views through this way, to decode out charge-discriminative words. 

As shown in Fig. \ref{model}, we further embed the charge label $\mathrm{v}$ to highlight the computing of hidden state $\mathbf{s}_t$ at time $t$ and $\mathbf{s}_t$ is merged as follows:
\begin{align}
\nonumber \mathbf{s}_t & = \mathrm{LSTM}_d(y_{t-1}, \mathbf{s}_{t-1}^{{v}}) \\
\nonumber \mathbf{s}_{t-1}^{{v}} & = f_v(\mathbf{s}_{t-1}, \mathrm{v}) \\ 
f_v & = \tanh(\mathbf{W}^{v} [\mathbf{s}_{t-1};\mathbf{E}^{{v}}_{[\mathrm{v}]}] + \mathbf{b}^{{v}}) 
\label{hidden}
\end{align}
where $\mathbf{W}^{{v}}$ and $\mathbf{b}^{{v}}$ are learnable parameters. In this way, the information of charge label can be embedded into $\mathbf{s}_t$. From Eq. (\ref{att-1}) and Eq. (\ref{att-2}), attention weights $\mathbf{c}_t$ are computed from $\mathbf{s}_t$, so encoding the charge label $\mathrm{v}$ to hidden states will make the model concentrate more on charge-related information from fact descriptions to help generate more accurate fact details.

\subsection{Model Training and Inference}
Suppose we are given the training data: $\{\mathrm{x}^{(i)}, \mathrm{y}^{(i)}, \mathrm{v}^{(i)}\}_{i=1}^N$, we aim to maximize the log-likelihood of generated rationales in court views given the fact descriptions and charge labels, so the loss function is computed as follows:
\begin{align}
\nonumber \mathcal{L(\theta)} & = -\sum_{i=1}^N \log P(\mathrm{y}^{(i)} | \mathrm{x}^{(i)}, \mathrm{v}^{(i)};\theta)  \\
 \nonumber  & = -\sum_{i=1}^N \sum_{j=1}^{|\mathrm{y}^{(i)}|} \log P(y_{j}^{(i)} | y_{ <j}^{(i)}, \mathrm{x}^{(i)}, \mathrm{v}^{(i)}; \theta)
\end{align}
We split the training data into multiple batches with size of $\mathrm{64}$ and adopt adam learning \cite{adam} to update the parameters in every batch data. At the inference time, we encode the fact descriptions and charge labels into vectors and use the decoder to generate rationales in court views based on Eq. (\ref{all-define}). We adopt the algorithm of beam search to generate rationales. Beam search size is set to $5$. To make generation process stoppable, an indicator tag ``$<$/s$>$'' is added to the end of the rationales sequences, and when ``$<$/s$>$'' is generated the inference process will be terminated. The generated word sequential paths will be ranked and the one with largest value is selected as the final rationales in court view. 

\section{Experiments}
\subsection{Data Preparation}
Following \citet{luo2017}, we construct dataset from the published legal documents in China Judgements Online\footnote{http://wenshu.court.gov.cn}. We extract the fact descriptions, rationales in court views and charge labels using regular expressions. The paragraph started with ``经审理查明'' (``our court identified that'') is regarded as the fact description and the part between ``本院认为'' (``our court hold that'') and the charge are regarded as the rationales. Nearly all the samples in dataset match this extraction pattern. Length threshold of $\mathrm{256}$ is set up, and fact description longer than that will be stripped, leaving too long facts for future study. We use the tokens of ``$<$$\text{name}$$>$", ``$<$$\text{num}$$>$" and ``$<$$\text{date}$$>$'' to replace the names, numbers and dates appearing in the corpus. We tokenize the Chinese texts with the open source tool of HanLP\footnote{https://github.com/hankcs/HanLP}. For charge labels, we select the top $50$ charge labels ranked by occurrences and leave the left charges as others. Details about our dataset are shown in Table \ref{dataset}.
\begin{table}
  \centering
 \small
  \begin{tabular}{l r}
  \toprule[0.8pt] 
  \# Training set &  $\mathrm{153706}$  \\ 
  \# Dev set &  $\mathrm{9152}$ \\ 
  \# Test set & $\mathrm{9123}$ \\ \hline
   Avg. \# tokens in fact desc. & $\mathrm{219.9}$ \\ 
   Avg. \# tokens in rationales & $\mathrm{30.6}$ \\ \hline
   Num. of  \# charge labels  & $\mathrm{51}$ \\ \hline
  \# Dict. size in fact desc. & $\mathrm{222482}$ \\
  \# Dict. size in rationales & $\mathrm{21305}$ \\
  \bottomrule[0.8pt]  
  \end{tabular}
  \caption{Statistics of our dataset.}\label{dataset}
\end{table}

For cases with multiple charges and multiple defendants, we can separate the fact descriptions and the court views according to the charges or the defendants. In this work, we only focus on the cases with one defendant and one charge, leaving the complex cases for future study, so we can collect large enough data from the published legal documents without human to annotate the data.


\subsection{Experimental Settings}
Word embeddings are randomly initialized and updated in the training process, with the size of $\mathrm{512}$ tuned from $\{\mathrm{256}, \mathrm{512}, \mathrm{1024}\}$. Charge label vectors are initialized randomly with size of $\mathrm{512}$. Maximal vocabulary size of encoder is set to $\mathrm{100}$K words and decoder is $50$K by stripping words exceeding the bounds. Maximal source length is $\mathrm{256}$ and target is $\mathrm{50}$. The hidden size of LSTM is $\mathrm{1024}$ tuned from $\{\mathrm{256}, \mathrm{512}, \mathrm{1024}\}$. We choose perplexity as the update metric. Early stopping mechanism is applied to train the model. The initial learning rate is set to $\mathrm{0.0003}$ and the reduce factor is $\mathrm{0.5}$. Model performance will be checked on the validation set after every $\mathrm{1000}$ batches training and keep the parameters with lowest perplexity. Training process will be terminated if model performance is not improved for successive $\mathrm{8}$ times. 

\subsection{Comparisons with Baselines}
\noindent{\textbf{Evaluation Metrics.}}
We adopt both automatic evaluation and human judgement for model evaluation. BLEU-4 score \cite{BLEU} and variant Rouge scores \cite{ROUGE} are adopted for automatic evaluation which have been widely used in many NLG tasks. 
We set up two evaluation dimensions for human judgement: 1) how \emph{fluent} of the rationales in court view is; 2) how \emph{accurate} of the rationales is, aiming to evaluate how many fact details have been accurately expressed in the generated rationales. We adopt $\mathrm{5}$ scales for both \emph{fluent} and \emph{accurate} evaluation ($\mathrm{5}$ is for the best). We ask three annotators who knows well about our task to conduct the human judgement. We randomly select $\mathrm{100}$ generated rationales in court views for every evaluated method. The three raters are also asked to judge whether rationales can be adopted for use in comprehensive evaluation (\emph{adoptable}) and record the number of adoptable rationales for every evaluated method. 

\begin{table}
  \centering
  \small
  \begin{tabular}{{p}{2.1cm}  p{0.8cm}<{\centering}  p{0.8cm}<{\centering} p{0.8cm}<{\centering} p{0.8cm}<{\centering}}
  \toprule[0.8pt] 
  \multicolumn{5}{r}{{\scshape{Automatic evaluation}}} \\ 
   \scshape{Model} (\%)  &  \scshape{B-4} & \scshape{R-1} & \scshape{R-2} & \scshape{R-L} \\ \hline
   \end{tabular}
   \begin{tabular}{{p}{2.1cm} p{0.8cm}<{\centering}  p{0.8cm}<{\centering} p{0.8cm}<{\centering} p{0.8cm}<{\centering}}
   $\text{Rand}_{\text{all}}$   & $\mathrm{6.4}$ & $\mathrm{26.5}$ & $\mathrm{6.2}$ & $\mathrm{25.1}$ \\
   $\text{Rand}_{\text{charge}}$  & $\mathrm{24.9}$ & $\mathrm{53.6}$ & $\mathrm{29.1}$ & $\mathrm{49.3}$ \\
   $\text{BM25}_{\text{f2f}}$  & $\mathrm{40.1}$ & $\mathrm{63.5}$ & $\mathrm{43.7}$ & $\mathrm{60.3}$ \\
   $\text{BM25}_{\text{f2f+charge}}$  & $\mathrm{42.8}$ & $\mathrm{67.1}$ & $\mathrm{47.4}$ & $\mathrm{63.8}$  \\
   MOSES+  &  $\mathrm{6.2}$ & $\mathrm{39.8}$ & $\mathrm{20.8}$ & $\mathrm{18.6}$ \\
   NN-S2S  &  $\mathrm{38.4}$ & $\mathrm{65.5}$ & $\mathrm{45.1}$ & $\mathrm{62.2}$  \\ 
   ${\text{RAS}}^\dagger$ & $\mathrm{44.1}^{\ast\ast}$ & ${\mathrm{69.1}^{\ast\ast}}$  & $\mathrm{50.3}^{\ast\ast}$ & $\mathrm{65.9}^{\ast\ast}$ \\
   \toprule[0.8pt]
  \end{tabular}
  \begin{tabular}{{p}{2.1cm} p{0.8cm}<{\centering}  p{0.8cm}<{\centering} p{0.8cm}<{\centering} p{0.8cm}<{\centering}}
   Ours  & $\mathbf{45.8}$ & $\mathbf{70.9}$ & $\mathbf{52.5}$  & $\mathbf{67.7}$    \\ 
    \toprule[0.8pt]
   \multicolumn{5}{r}{{\scshape{Human judgement}}} 
   \\ 
 \end{tabular}
  \begin{tabular}{{p}{2.2cm}  p{1.1cm} p{1.1cm}  p{1.2cm}} 
   \scshape{Model}  &  \scshape{{fluent}} &  \scshape{acc.}  &  \scshape{adopt.}\tiny{($\%$)}  \\ \hline
   $\text{BM25}_{\text{f2f}}$ & $\mathrm{4.95}$ & $\mathrm{3.66}^{\ast\ast}$ &  $\mathrm{0.47}^{\ast\ast}$ \\ 
  $\text{BM25}_{\text{f2f+charge}}$  & ${4.94}$  & $\mathrm{3.90}^{\ast\ast}$  & $\mathrm{0.50}^{\ast\ast}$ \\ 
  MOSES+ & $\mathrm{1.39}^{\ast\ast}$ & $\mathrm{1.31}^{\ast\ast}$ & $\mathrm{0}^{\ast\ast}$ \\
  NN-S2S & $\mathrm{4.97}$ & $\mathrm{4.07}^{\ast\ast}$ & $\mathrm{0.62}^{\ast}$ \\ 
  $\text{RAS}^\dagger$ & $\mathrm{4.96}$ & $\mathrm{4.25}^{\ast}$ & $\mathrm{0.64}^{\ast}$ \\ \toprule[0.8pt]
  Ours   & $\mathrm{4.93}$ & ${\mathbf{4.54}}$  & $\mathbf{0.72}$  \\ 
  \toprule[0.8pt]
  \end{tabular}
  \caption{\small{Results of automatic evaluation and human judgement with BLEU-4 and full length of F1 scores of variant Rouges. Best results are labeled as boldface. Statistical significance is indicated with $\ast\ast$($p < 0.01$) and $\ast$ ($p < 0.05$) comparing to our full model.}} \label{automatic-eval-result}
\end{table}

\noindent{\textbf{Baselines.}}

$\bullet$ \textbf{Rand} is to randomly select rationales in court views from the training set (method of \textbf{$\text{Rand}_{\text{all}}$}). We also randomly choose rationales from pools with same charge labels (\textbf{$\text{Rand}_{\text{charge}}$}). Adopting Rand method is to indicate the low bound performance of $\text{\scshape{Court-View-Gen}}$. 


$\bullet$ $\textbf{BM25}$ is a retrieval baseline to index the fact description match to the input fact description with highest BM25 score \cite{BM25} from the training set, and use its rationales as the result ($\textbf{BM25}_{\textbf{f2f}}$). Similar fact descriptions may have the similar rationales. 
Fact descriptions from pools with same charges are also retrieved ($\textbf{BM25}_{\textbf{f2f+charge}}$), to see how much improvement that adding charge labels can gender.

$\bullet$ \textbf{MOSES+} \cite{MOSES} is a phrase-based statistical machine translation system mapping fact descriptions to rationales. KenLM \cite{KenLM} is adopted to train a trigram language model on the target corpus of training set which is tuned on the validation set with MERT.

$\bullet$ \textbf{NN-S2S} is the basic Seq2Seq model without attention from \citet{Seq2SeqNN} for machine translation. We set one LSTM layer for encoding and another one LSTM layer for decoding. We adopt perplexity for training metric and select the model with lowest perplexity on validation set. 

$\bullet$ $\textbf{RAS}^\dagger$ is an attention based abstract summarization model from \citet{Sum-Chopra}. To deal with the much longer fact descriptions, we exploit the more advanced bidirectional LSTM model for the encoder instead of the simple convolutional model. Another LSTM model is set as the decoder coherent to \citet{Sum-Chopra}. 


\noindent{\textbf{Experimental Results.}}
In automatic evaluation from Table \ref{automatic-eval-result}, the evaluation scores are relatively high even for method of $\text{Rand}_{\text{charge}}$, which indicates that the expressions of the rationales with same charge labels are similar with many overlapped n-grams, such that the rationales for \emph{crime of theft} usually begin with ``以非法占有为目的'' (``in intention of illegal possession").\
Accurately generating fact details like degree of injury or time of theft is more difficult. 
Retrieval method by adding charge labels is the strong baseline even better than basic Seq2Seq model.\ Adding attention mechanism will improve the performance indicated by the method of $\text{RAS}^\dagger$  which is superior to retrieval methods. By exploiting charge labels, our full model achieves the best performance. The performances of statistical machine translation model are really poor, for it requiring the lengths of parallel corpus to be similar. 

In human evaluation, we can see that retrieval methods can not accurately express fact details, for that it is hard to retrieve rationales containing details all matching the fact descriptions. However, our system can learn to generate fact details by analyzing fact descriptions. Dropping attention mechanism will have negative effects on model performance. $\text{RAS}^\dagger$ has worse performance in $\text{\scshape{acc.}}$ whose main reason may lie in that $\text{RAS}^\dagger$ can not generate charge-discriminative rationales with \emph{deduced information}, which demonstrates that our task is not the simple $\text{\scshape{doc-sum}}$ task.  
For the \emph{fluent} evaluation, generation models are highly close to retrieval methods whose rationales are written by humans, which reflects that the generation models can generate highly natural rationales.


\subsection{Further Analysis}
\noindent{\textbf{Impact of Exploiting Charge Labels.}}

\begin{figure}
\centering\includegraphics[width = \columnwidth]{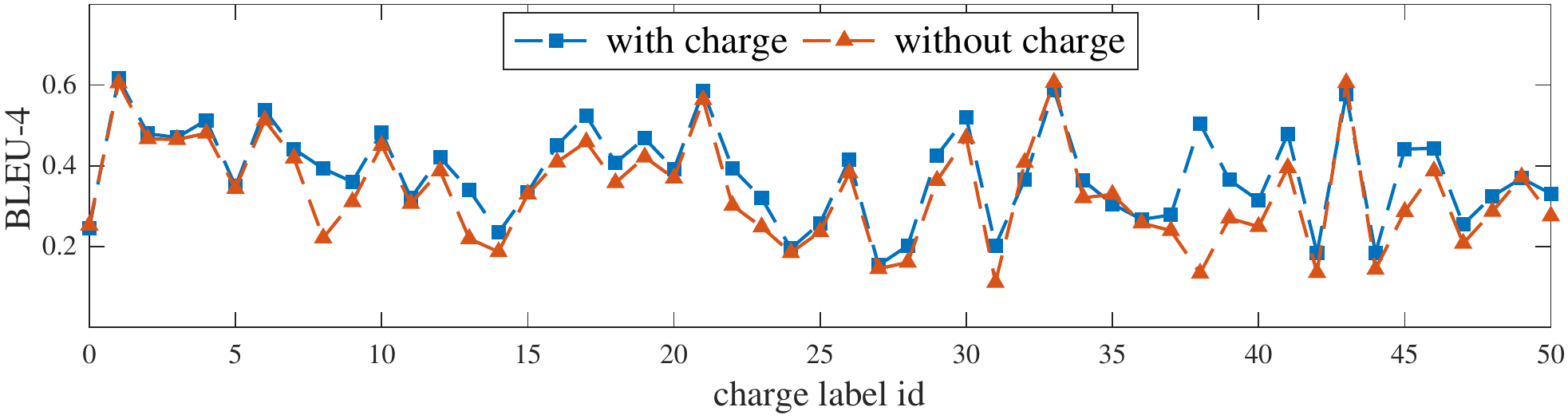}
\caption{\small{Results of impact of exploiting charge labels evaluated charge to charge in the metric of BLEU-4 (similar results can gender in other three metrics but are omitted for space saving).}}
\label{label-all-charge}
\end{figure}

$\bullet$ \textbf{Charge2Charge Analysis.} 
We first analyze the effects of exploiting charge labels on model performance charge to charge, by dropping to encode charges based on our full model. 
From the results shown in Fig. \ref{label-all-charge}, we can find that the results can be improved much by exploiting charge labels among nearly all charges. This result also indicates that the non-distinct fact descriptions are common among nearly all charges and reflects the difficulty of this task, but utilizing charge labels can release the seriousness of the problem. 


$\bullet$ \textbf{Charge-discriminations Analysis.} We further evaluate the effects of charge labels for charge-discriminations improvement on specific charges with non-distinct fact descriptions: \emph{intentional homicide}, \emph{negligent homicide}, \emph{duty embezzlement} and \emph{corruption}. 
For every charge, two participants are asked to count the number of rationales that are relevant to the charge on $\mathrm{20}$ randomly selected candidates. 

\begin{figure}
\centering\includegraphics[width = \columnwidth]{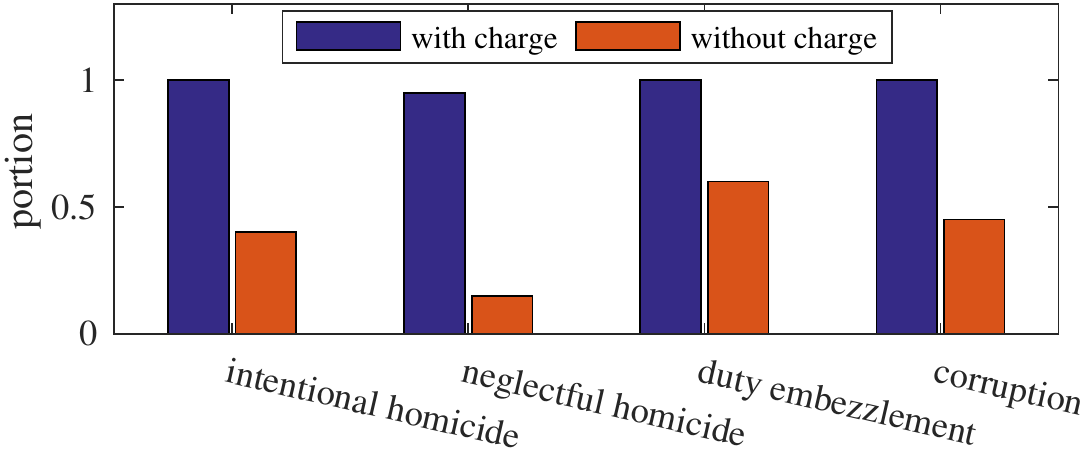}
\caption{\small{Portions of charge-discriminative rationales in court views for every charge with $\mathrm{20}$ candidates.}}
\label{class-relevance}
\end{figure}

From Fig. \ref{class-relevance}, the number of charge discriminative rationales can be much improved among every charge by utilizing charge information, which 
demonstrates that charge labels can provide with much extra charge-related information to deal with latent information in fact descriptions. For crimes of \emph{homicide}, the motives for killing are latent in the descriptions of killing without direct statement, but our system can learn to align the motives in rationales to the charge labels which are the strong distinct indicator for the two motives. 

\noindent{\textbf{Ablation Study.}} We also ablate our full model to reveal different components of encoding charge labels for performance improvement. As shown in Table \ref{tab:ablation}, `` / softmax comp.'' is to remove the part in Eq. (\ref{dict}) and yields worse performance than our full model, but better than `` / charge comp.'' that ignores to encode charge labels, which is same to the situation of `` / hidden comp.'' that removes the part in Eq. (\ref{hidden}). Our full model is still better than the ablated models. This finding shows that both of the methods of exploiting charge labels can improve model performance and stacking them will achieve better results. 
\begin{table}
  \centering
  \small
  \begin{tabular}{{p}{2.5cm}  p{0.75cm}<{\centering}  p{0.75cm}<{\centering} p{0.75cm}<{\centering} p{0.75cm}<{\centering}}
  \toprule[0.8pt] 
  \multicolumn{5}{r}{{\scshape{Ablation Study}}} \\ 
   \scshape{Model} (\%)  &  \scshape{B-4} & \scshape{R-1} & \scshape{R-2} & \scshape{R-L} \\ \hline
   \end{tabular}
   \begin{tabular}{{p}{2.5cm} p{0.75cm}<{\centering}  p{0.75cm}<{\centering} p{0.75cm}<{\centering} p{0.75cm}<{\centering}}
   $\text{Our System}$   & $\mathrm{45.8}^{\ast\ast}$ & $\mathrm{70.9}^{\ast\ast}$ & $\mathrm{52.5}^{\ast\ast}$ & $\mathrm{67.7}^{\ast\ast}$ \\
   
   $\text{\small{\ \ \ \ \ \ / softmax comp.}}$  & $\mathrm{45.7}^{\ast\ast}$ & $\mathrm{70.8}^{\ast\ast}$ & $\mathrm{52.3}^{\ast\ast}$ & $\mathrm{67.5}^{\ast\ast}$ \\
   
   $\text{\small{\ \ \ \ \ \ / hidden comp.}}$  & $\mathrm{45.7}^{\ast\ast}$ & $\mathrm{70.2}^{\ast}$ & $\mathrm{51.9}^{\ast}$ & $\mathrm{67.0}^{\ast}$ \\
   
   $\text{\small{\ \ \ \ \ \ / charge comp.}}$  & $\mathrm{43.7}$ & $\mathrm{68.6}$ & $\mathrm{49.7}$ & $\mathrm{65.5}$  \\
\bottomrule[0.8pt]
  \end{tabular}\caption{\small{Results of ablation study. Statistical significance is indicated with $\ast\ast$ ($p < 0.01$) and $\ast$ ($p < 0.05$) comparing to the ablation of `` / charge comp.''.}} \label{tab:ablation}
 \end{table}

\begin{figure}
\centering\includegraphics[width = 8.3cm]{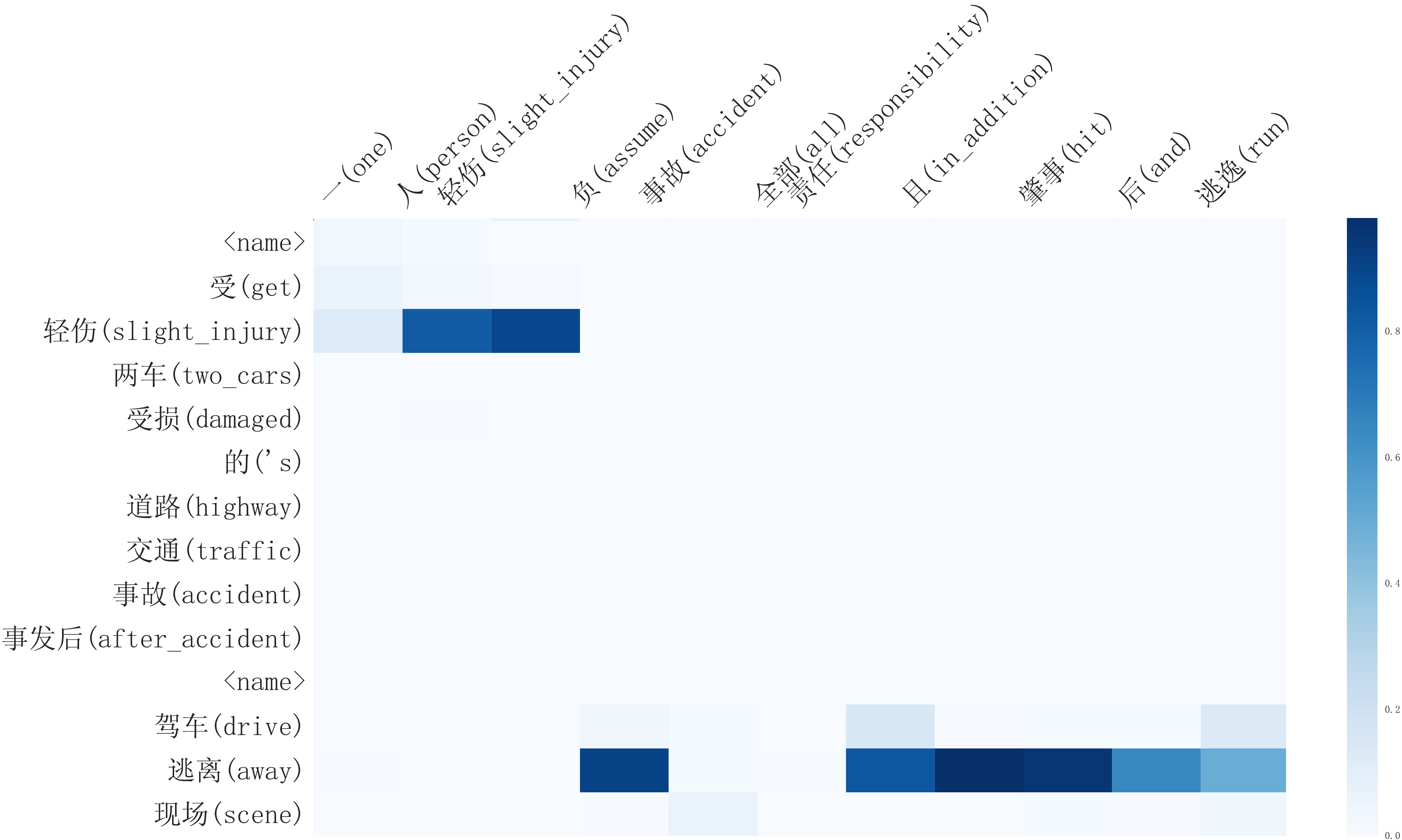}
\caption{\small{Heat map for attention mechanism analysis. The column is the source and the raw is the target.}}
\label{attention-map}
\end{figure}

\noindent{\textbf{Attention Mechanism Analysis.}} Heat map in Fig. \ref{attention-map} is used to illustrate the attention mechanism. The ``slight injury'' is aligned between the source and target. ``responsibility'' and ``run'' are well aligned to ``away'', which demonstrate the efficiency of attention mechanism for generating fact details by forcing context vectors to focus more on fact details. 

\begin{figure}
\centering\includegraphics[width = \columnwidth]{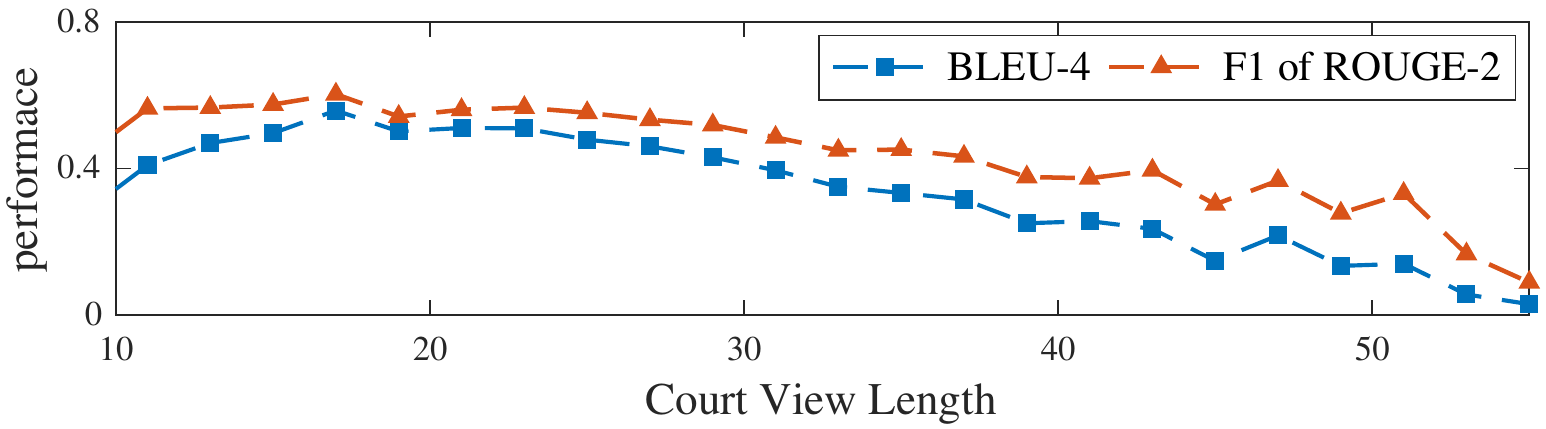}
\caption{\small{Model performance by rationales length with BLEU-4 and full length of F1 of Rouge-2.}}
\label{length}
\end{figure}

\noindent{\textbf{Performance by Reference Size.}} 
We further investigate the model performance by rationales length in court views. As shown in Fig. \ref{length}, not surprisingly the model performance drops when the length of reference rationales increases. Within the size of $\mathrm{30}$, BLEU-4 score can maintain around $\mathrm{0.4}$ and F1 score keeps around $\mathrm{0.5}$. Exceeding the length of $\mathrm{30}$, model performance decreases dramatically. 

\begin{figure}
\centering\includegraphics[width = \columnwidth]{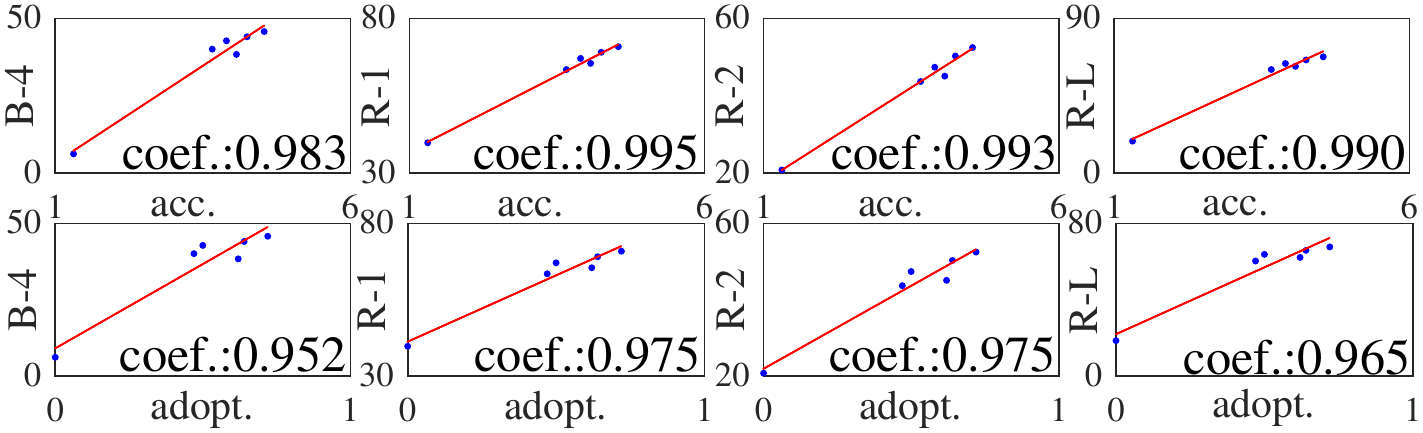}
\caption{\small{$\text{\scshape{acc.}}$ and $\text{\scshape{adopt.}}$ of human judgement predict automatic evaluation scores.}}
\label{fig:correlation}
\end{figure}

\noindent{\textbf{Human eval. vs. Automatic eval.}} \emph{Are BLEU and Rouge suitable for $\text{\scshape{Court-}}$$\text{\scshape{View-}}$$\text{\scshape{Gen}}$ evaluation}? Following the work of \cite{BLEU,human-vs-auto}, for the models evaluated in human judgemnet, we draw the linear regressions of their BLEU-4 and variant Rouge scores, as the function of $\text{\scshape{acc.}}$ and $\text{\scshape{adopt.}}$ from human judgement respectively as shown in Fig. \ref{fig:correlation}. From the results, we can find that automatic evaluations track well with the human judgement with high correlation coefficients. This finding demonstrates that BLEU-4 and variant Rouges are adoptable for $\text{\scshape{Court-}}$$\text{\scshape{View-}}$$\text{\scshape{Gen}}$ evaluation and provides the basis for future studies on this task. 

\noindent{\textbf{Error Analysis.}} Our model has the drawback of \emph{generating latent fact details}, which appear in rationales but are not clearly expressed in fact descriptions. For example, for the time of theft in charge of \emph{larceny}, the term of ``多次'' (``several times'') appears in rationales but may not be expressed in fact descriptions directly, only with descriptions of larceny but without exact term for this detail, so it will be hard for attention mechanism to learn to align ``多次'' in rationales to latent information in fact descriptions. In the generated rationales on test set, we find that only $\mathrm{42.4\%}$ samples can accurately extract out the term of ``多次''. It may need designed rules to deal with such details, like that count the time of theft from the descriptions, and if the time exceeds $\mathrm{1}$ then the term of ``多次'' can be generated in rationales.

\begin{figure*} 
\centering\includegraphics[width = \textwidth]{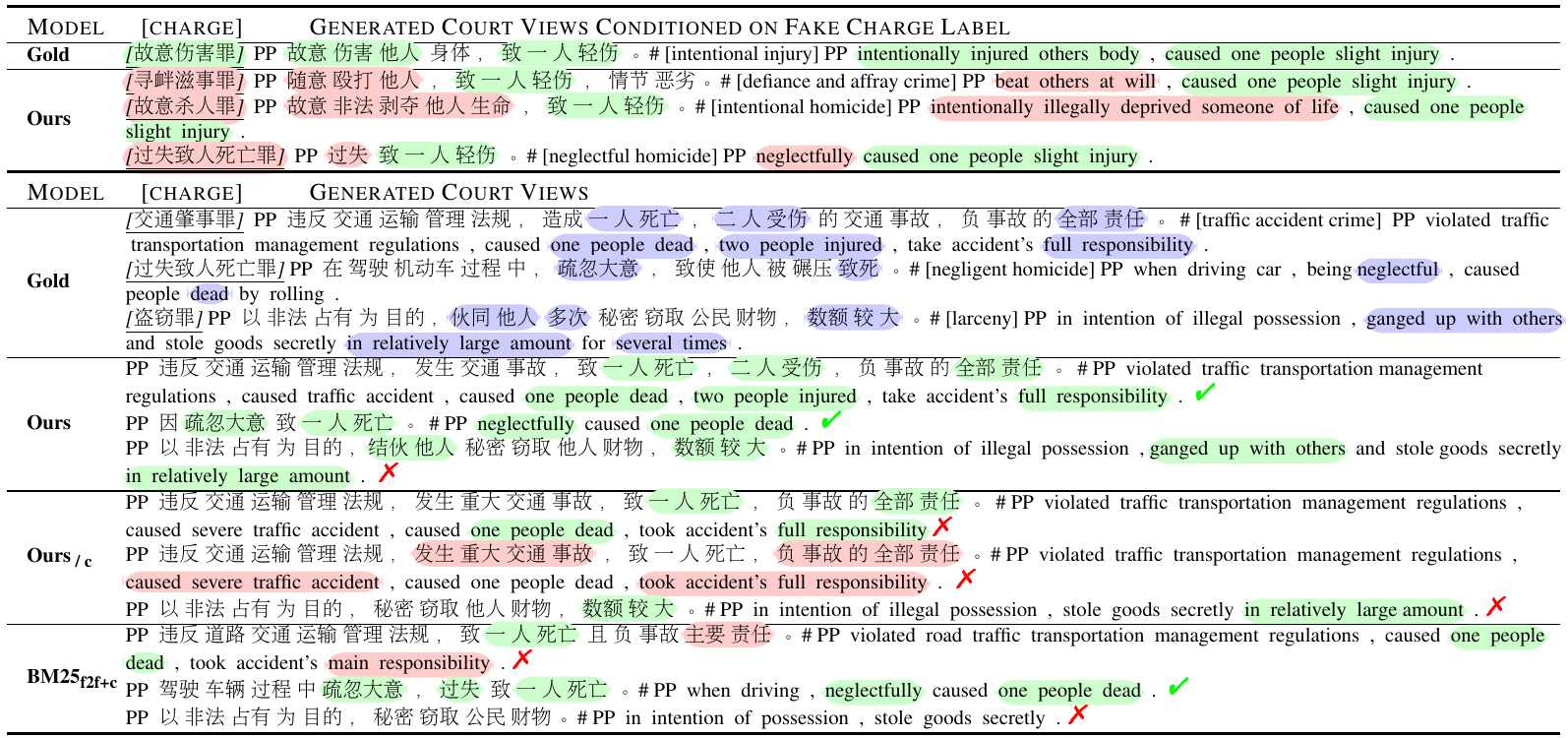}
\caption{\small{Fake charge label conditioned generated rationales in court views and examples of generated rationales.}}
\label{case-study}
\end{figure*}

\subsection{Analysis through Cases}
\noindent{\textbf{Fake Charge Label Conditioned Study.}} What generated rationales in court views will be if they are conditioned on fake charge labels?  We select one fact description with gold charge of \emph{intentional injury}, then generate rationales conditioned on fake charges of \emph{defiance and affray crime}, \emph{intentional homicide} and \emph{neglectful homicide}. 

From Fig. \ref{case-study}, the rationales conditioned on fake charges will be partly relevant to fake charge labels and also maintain fact details from the input fact description of gold charge. For the fake charge of \emph{intentional homicide}, its fact details should be ``caused someone dead'', but instead express ``causing someone slight injury'' which is relevant to charge of \emph{intentional injury}. For charge prediction systems, the discriminations between fact details and charges will help to remind people that the prediction results may be unreliable.

\noindent{\textbf{Case Study.}} Examples of generated rationales in court views are shown in Fig. \ref{case-study}. Generally speaking, our full label-conditioned model has high accuracy on generating fact details better than baseline models. For charges of \emph{traffic accident crime} and \emph{negligent homicide}, all fact details are generated. The extra information from charge labels helps the model to capture more important fact details, by forcing model to pay more attention to charge-related information in fact descriptions. 

As for the charge-discrimination analysis, from the rationales of \emph{negligent homicide}, we can infer that its fact description may relate to a traffic accident, which is non-distinct from that for \emph{traffic accident crime}. Without encoding charge labels, $\text{Ours}_{\text{ / c}}$ wrongly generates the rationales coherent to \emph{traffic accident crime}, because traffic accidents 
are the strong indicator for traffic crimes, but the charge label will provide extra bias towards the \emph{homicide crime}, so our full model can generate highly discriminative rationales. 
Utilizing charge labels, retrieval method can easily retrieve charge-related rationales, but hard to index rationales with accurate fact details. 
For charge of \emph{larceny}, our full model extracts nearly all fact details but misses the fact of ``多次''(``several times''), reflecting the shortcoming of dealing with latent details. 

\section{Conclusion and Future Work}
In this paper, we propose a novel task of court view generation and formulate it as a text-to-text NLG problem. We utilize charge labels to benefit the generation of charge-discriminative rationales in court views and propose a label-conditioned Seq2Seq model with attention for this task. Extensive experiments show the efficiency of our model and exploiting charge labels. 

In the future: 
1) More advanced technologies like reinforcement learning \cite{RL} can be introduced to generate latent fact details such as the time of theft more accurately; 
2) In this work, we only generate rationales in court views omitting charge prediction, it is interesting to see whether jointly generating the two parts will benefit both of the tasks;
3) Studying verification mechanism is meaningful to judge whether generated court views can really be adopted which is important for $\text{\scshape{court-view-gen}}$ in practice;
4) More complex cases with multiple charges and multiple defendants will be considered in the future. 

\section*{Acknowledgments}
Firstly, we would like to thank Yansong Feng, Yu Wu, Xiaojun Wan, Li Dong and Pengcheng Yin for their insightful comments and suggestions. We also very appreciate the comments from anonymous reviewers which will help further improve our work. This work is supported by National Natural Science Foundation of China (No. 61602490) and National Key R\&D Plan (No. 2017YFB1402403). The work was done when Hai Ye interned in Beihang University from August, 2017 to January, 2018.

\bibliographystyle{acl_natbib}

\end{CJK*}
\end{document}